\documentclass{article}
\usepackage{ijcai19}

\usepackage{times}
\usepackage{soul}
\usepackage{url}
\usepackage[hidelinks]{hyperref}
\usepackage[utf8]{inputenc}
\usepackage[small]{caption}
\usepackage{subcaption}
\usepackage{graphicx}
\usepackage{amsmath}
\usepackage{booktabs}
\usepackage{algorithm}
\usepackage{algorithmic}
\usepackage{listings}
\usepackage{tikz}
\usepackage{siunitx}

\newcommand{\ie}{\textit{i}.\textit{e}. }
\newcommand{\eg}{\textit{e}.\textit{g}. }
\usepackage{xcolor}

\tikzset{
  square arrow/.style={
    to path={-- ++(-10,-.25) -| (\tikztotarget)}
  }
}

\title{Injecting Prior Knowledge for Transfer Learning into\\ Reinforcement Learning Algorithms using Logic Tensor Networks}

\author{
Samy Badreddine$^1$
\and
Michael Spranger$^2$
\affiliations
$^1$Universit{\'e} Libre de Bruxelles\\
$^2$Sony Computer Science Laboratories Inc.
\emails
samy.badreddine@gmail.com,
michael.spranger@gmail.com
}

\begin{document}

\maketitle

\begin{abstract}
  Human ability at solving complex tasks is helped by priors on object and event semantics of their environment. This paper investigates the use of similar prior knowledge for transfer learning in Reinforcement Learning agents. In particular, the paper proposes to use a first-order-logic language grounded in deep neural networks to represent facts about objects and their semantics in the real world. Facts are provided as background knowledge a priori to learning a policy for how to act in the world. The priors are injected with the conventional input in a single agent architecture. As proof-of-concept, the paper tests the system in simple experiments that show the importance of symbolic abstraction and flexible fact derivation. The paper shows that the proposed system can learn to take advantage of both the symbolic layer and the image layer in a single decision selection module.
\end{abstract}

\section{Introduction}
\begin{figure}
    \centering
    \includegraphics[width=\columnwidth]{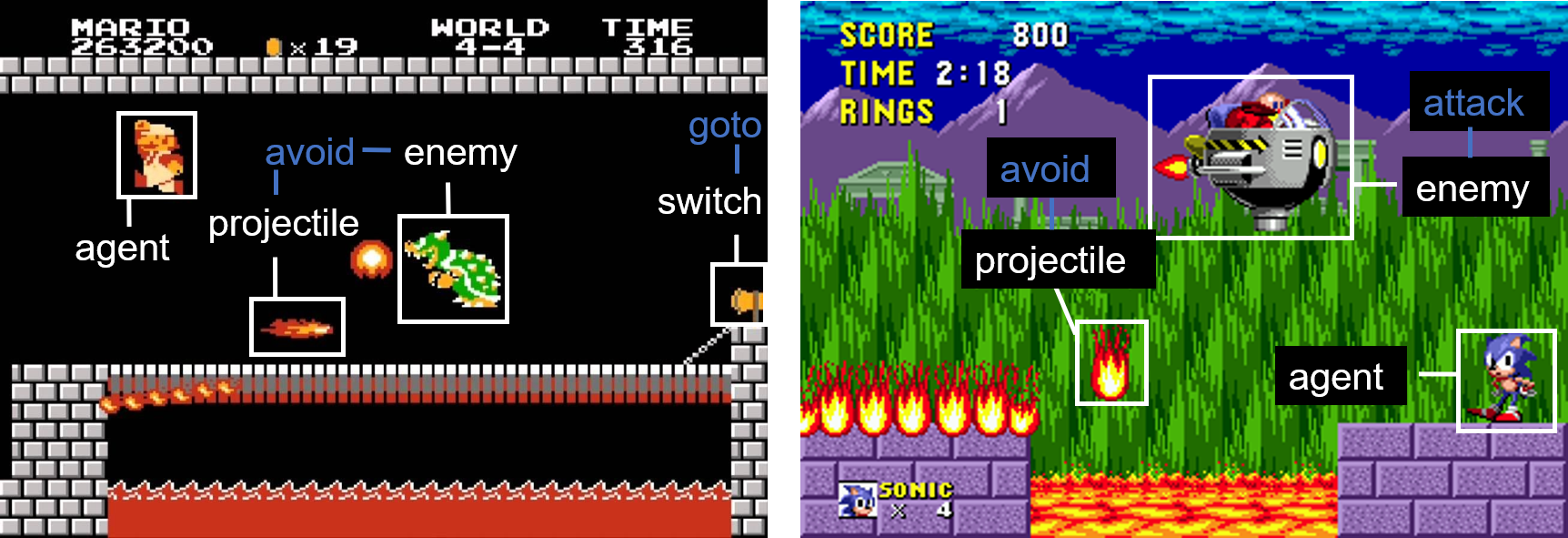}
    \caption{Left: Super Mario Bros. (1985) - Right: Sonic the Hedgehog (1991). Humans can easily identify objects semantics in both environments (enemy, projectile, agent). Players with past experience in these games also know facts such as: the agent should avoid projectiles, the Mario boss is defeated by activating a trap, the Sonic boss must be attacked upfront. The paper investigates the injection of such priors on object semantics and facts in DRL.}
    \label{fig:game_priors}
\end{figure}

Recently, much of AI and ML is concerned with end-to-end learning for solving various complex tasks such as Atari Games and Go~\cite{mnih_human-level_2015,silver2017mastering}. Almost all recent progress in this field has been driven by applications of deep neural networks to Reinforcement Learning (RL) tasks -- a subfield known as Deep Reinforcement Learning (DRL). While DRL has shown important progress over the past years, state-of-the-art algorithms still struggle to achieve human-like performance within limited training time. Traditional DRL agents' learning is slow and algorithms require lots of training data. Humans on the other hand are able to quickly understand and solve tasks they have never seen before in complex environments. It is likely that a large part of the explanation for such efficiency lies in the usage of prior knowledge. For example, experiments with human participants \cite{dubey_investigating_2018} show that humans are able to solve complex tasks using specific priors whereas humans fail if such priors are not applicable. Such priors are illustrated in Figure \ref{fig:game_priors}.

Inspired by human cognition, this paper proposes to exploit prior knowledge for transfer learning in DRL agents. More specifically, we focus on {\it object semantics priors} and describe high-level symbolic facts about objects in the environment --\eg object $x$ is an enemy, $y$ a key, $z$ a door, etc.-- using a first-order language. The knowledge is provided by the human (as prior knowledge) and joined to the image describing the environment. A DRL algorithm is then trained on conjoint image and semantic data and can choose to exploit prior information if it helps performance and learning. That is the system proposed in this paper can learn to take advantage of both the symbolic layer and the conventional layer in a single decision selection module. Also because knowledge is provided in a first-order language, the system is easily extended with new facts and relationships about objects and the environment. We test our framework in a simple grid-world environment. 

The system presented in this paper relies on Logic Tensor Networks (LTN) \cite{serafini_logic_2016} for representing prior knowledge. LTN has been applied to image segmentation and interpretation \cite{donadello_logic_2017} and also hierarchy learning. So far LTN has not been applied to Reinforcement Learning. To the best of our knowledge this paper presents the first attempt at applying a first-order, neural network grounded system (such as LTN) to Reinforcement Learning.

The paper proceeds by discussing related work, we then illustrate the target environments followed by a detailed description of the proposed framework and system. Following results, we discuss some of the implications of our approach and future work.

\section{Related Work}
The combination of Deep Neural Networks and Reinforcement Learning has been the most successful approach to Reinforcement Learning in the last ten years. Most famously DRL has solved ATARI games \cite{mnih_human-level_2015} and the game of Go \cite{silver2017mastering}. ATARI and Go both are essentially discrete state and action space Markov Decision Problems. But, DRL has also been applied to continuous control problems with multiple proposals existing for continuous state and action spaces \cite{duan2016benchmarking,lillicrap2015continuous,mnih2016asynchronous}. In principle our approach is compatible with all discrete time Reinforcement Learning problems. Our approach combines rich (continuous) input from images or other sensori information with symbolic information and we then apply any DRL learning system. Our overall system differs from pure DRL systems by being able to easily incorporate prior and background knowledge available in a first-order grounded language.


All of the approaches mentioned in the previous paragraph are examples of end-to-end trainable systems (in the case of DRL). DRL takes as input raw images or sensor data and outputs discrete or continuous actions. It is difficult to add prior knowledge to such systems in a systematic manner. Similarly, traditional relational RL approaches do not take into account the grounding in sensorimotor spaces and how to learn in those spaces conjointly with the symbolic information. 

How to combine such systems effectively has led recently to new work on how to use traditional symbolic knowledge representations in Reinforcement Learning. For instance, some recent work introduces symbolic front ends on top of neural back ends \cite{garnelo_towards_2016}.  The neural back-end is responsible of conceptual abstraction from the image and maps the raw input to symbolic representations. A symbolic layer then represents information in separate streams for each symbol, before a decision module aggregates them using heuristics. Others have tried to add common sense priors in the heuristic aggregation~\cite{garcez2018towards}. The system presented in this paper integrates different sources of information in a single representation before a DRL algorithm can learn to make choices using either symbolic information or raw pixel data. 

Another recent example of integrating symbolic information in Reinforcement Learning is Bougie et al. \shortcite{bougie2018combining}. They propose two streams of processing, one where the symbolic information is processed and one where the pixel data is used. In the end there are two actions and a supervised learning module selects which action to take either the symbolic or DRL output. Our architecture presented in this paper simplifies into a single action selection stream and can learn to take advantage of both the symbolic layer and the image layer in a single decision selection module.

\section{Experimental Setup and Task}
\begin{figure}
    \centering
    \includegraphics[width=0.5\columnwidth]{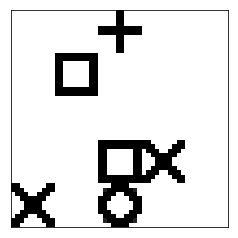}
    \caption{Simple game environment. The agent is represented by the $+$ sign. It acts by moving on a $5 \times 5$ grid and collects objects of different types \emph{circle}, \emph{square} and \emph{cross}. Scenarios set the object types as being \emph{target} (+1 reward on collection) or \emph{avoid} (-1 reward on collection).}
    \label{fig:exp_setup}
\end{figure}
\label{sec:experiments}
We test our system in a simple game environment inspired by previous work \cite{garnelo_towards_2016}. The environment consists of different types of objects and an agent (see  Figure \ref{fig:exp_setup}). Objects differ in shape: \emph{circle}, \emph{square} and \emph{cross}. 

In each environment objects of all three types are present. The task for the agent is to collect all objects of a particular \emph{target type} -- \eg all circles -- and avoid all objects of \emph{avoid type} -- \eg squares --. The agent receives a +1 reward upon collecting a \emph{target type} object, a -1 reward upon collecting an \emph{avoid type} object, a zero reward in all other cases.

The environment is represented by a $50\times 50$ image with objects of size $10\times 10$ in $5\times 5$ cells. The agent can move in the environment with the following actions: \emph{move-up}, \emph{move-down}, \emph{move-left}, \emph{move-right}. An object is collected when the agent enters a field. 

Number and position of objects as well as the position of the agent are randomized for each trial. A trial ends when the agent has collected all objects associated with a positive reward or after 50 steps.

From this basic environment structure with 3 types of objects and 1 agent, we create two experiments: \emph{Experiment I - Symbolic abstraction} and \emph{Experiment II - Fact derivation}

\subsection*{Experiment I - Symbolic abstraction}
The agent is trained on the same game but rendered with different colors.
\begin{description}
\item[Setting 1] black objects, white background
\item[Setting 2] white objects, black background
\item[Setting 3] red objects, blue background
\item[Setting 4] blue objects, red background
\end{description}
The different settings represent separate video games where objects and backgrounds look visually different but have the same meaning -- \eg enemies, hero, etc.--. The goal of this experiment is to highlight how a symbolic layer helps to transfer collection/avoidance strategies across object properties (a circle is a circle if drawn in red or in white).

\subsection*{Experiment II - Fact derivation}
The types of objects that have to be collected change over time.

\begin{description}
\item[Scenario 1] target circle (+1), avoid cross (-1)
\item[Scenario 2] target cross (+1), avoid circle (-1)
\item[Scenario 3] target square (+1), avoid cross (-1)
\item[Scenario 4] target cross (+1), avoid square and circle (-1)
\end{description}
The scenarios represent different video games where objects are similar in type but interaction is different. The goal of this experiment is to highlight how a symbolic layer can help to transfer (collection/avoidance) strategies across object types.

\begin{figure}
    \centering
    \includegraphics[width=1.\columnwidth]{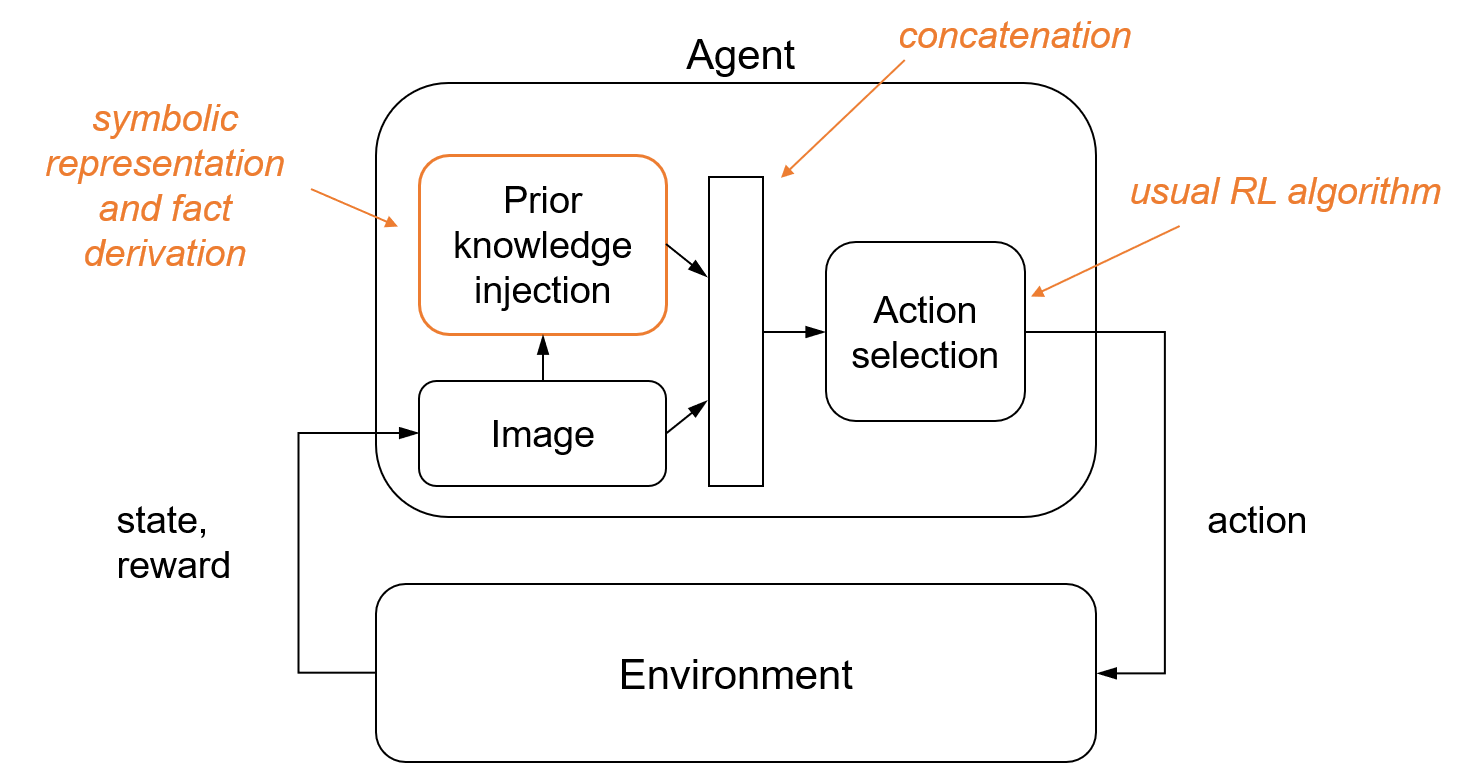}
    \caption{Architecture overview: the prior knowledge derived input is injected conjointly with the original input in the action selection algorithm}
    \label{fig:architecture-overview}
        \centering
    \includegraphics[width=1.\columnwidth]{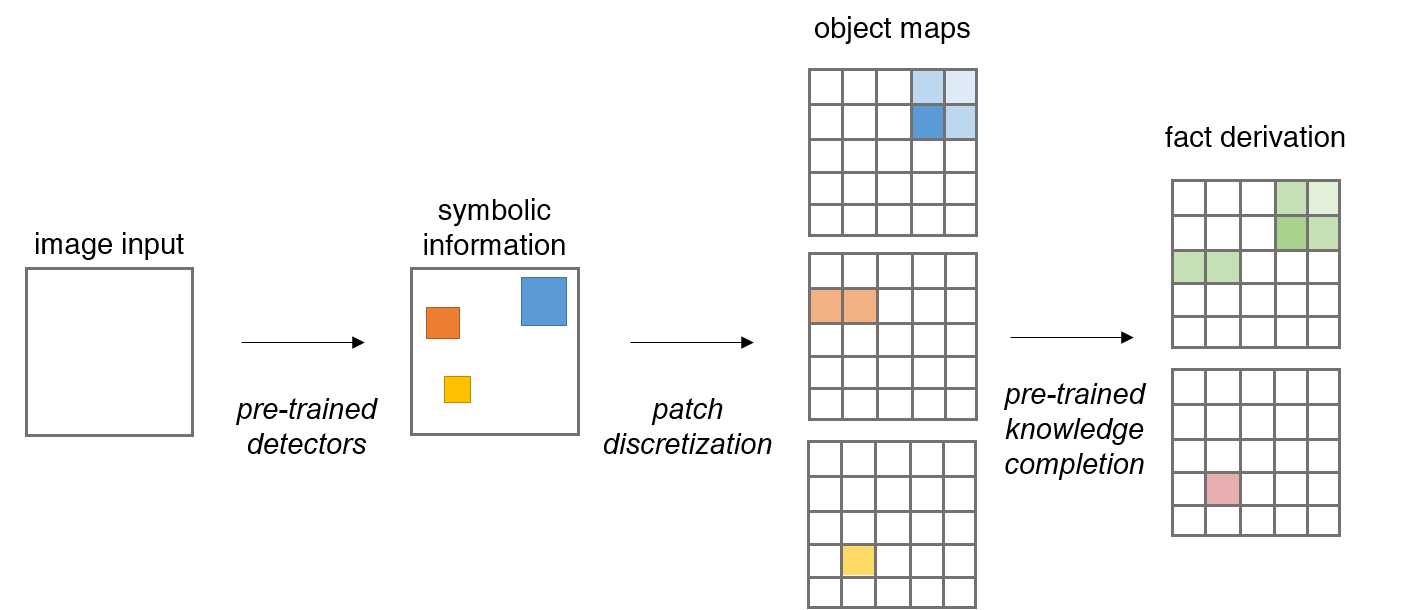}
    \caption{Prior knowledge injection process}
    \label{fig:architecture-prior-knowledge-injection}

\end{figure}

\section{Architecture Overview}

Our task setting is a sequential task and it can be principally solved using Reinforcement Learning (RL). The basic idea for RL is having an agent trying to solve a task by observing the environment through its sensors, choosing to act accordingly and occasionally receiving a reward for its actions. The goal of the agent is to learn a policy that maximizes expected reward across trials. We propose the following agent architecture to solve tasks such as the one discussed here (see also Figure \ref{fig:architecture-overview}). Overall our architecture consists of the following parts.

\begin{description}
\item[Input] The current image of the environment. Here we use a 50x50 pixel image.
\item[Prior Knowledge Injection] The image is processed with prior knowledge. Derived representations similar to feature maps are computed and augment the raw pixel information.
\item[Concatenation] Both the original input and the prior knowledge derived maps are combined into a single representation. 
\item[Action Selection] The concatenation result is the input for a Deep Reinforcement Learning algorithm. For the environment discussed in this paper, algorithms that can deal with image-like input spaces and discrete action spaces are appropriate. Here, we apply a Double Dueling Deep Q Network architecture with experience replay. 
\item[Output] The output is one of 4 possible actions (\emph{move-up}, \emph{move-down}, \emph{move-left}, \emph{move-right})
\end{description}

The following Sections give more details on the key parts of the architecture and how they interact.

\section{Prior Knowledge Representation with Logic Tensor Networks}
We inject prior knowledge into the agent using a three step process (Figure \ref{fig:architecture-prior-knowledge-injection}).

\begin{description}
\item[Symbolic descriptions] Symbolic reasoning on objects of an image is only possible if high-level symbolic descriptions are available to the agent. We extract these descriptions from the low-level pixel features in each image using pre-trained object detectors. We have classifiers for $\mathrm{agent},\mathrm{circle},\mathrm{square},\mathrm{cross}$.
\item[Object maps] We discretize the environment in a patch representation similar to feature maps. Being a grid, our game environment is easily described with $5 \times 5$ maps. Each channel of the maps represents an object type predicate and is filled with the detection results of the previous step. Each map is filled with the detection results of the previous step, using interpolation methods if the dimensions mismatch.
\item[Fact derivation] Facts about which objects to {\it avoid} and which to {\it goto} are computed based on background knowledge (axioms) for the currently active scenario.
\end{description}

This leads to the following grounded theory.

\begin{description}
\item[Domain] We assume a domain of objects $\mathcal{O}$ that consists in each cell of the object maps -- that is, the image patches of size $10 \times 10$.
\item[Predicates] We consider the following type predicates $\mathrm{circle},\mathrm{square},\mathrm{square},\mathrm{agent}$ and the following more general predicates $\mathrm{avoid},\mathrm{goto}$.
\item[Axioms] For each scenario we provide background knowledge in the form of axioms about which objects to avoid and which to go to. For instance, in \emph{Scenario 1} the following axioms hold and are used:
\begin{eqnarray}
    \forall x \in \mathcal{O}: \mathrm{circle}(x) &\leftrightarrow& \mathrm{goto}(x)\\
    \forall x \in \mathcal{O}:  \mathrm{cross}(x) &\leftrightarrow& \mathrm{avoid}(x)
\end{eqnarray}
\end{description}

\begin{figure}
    \centering
    \includegraphics[width=1.\columnwidth]{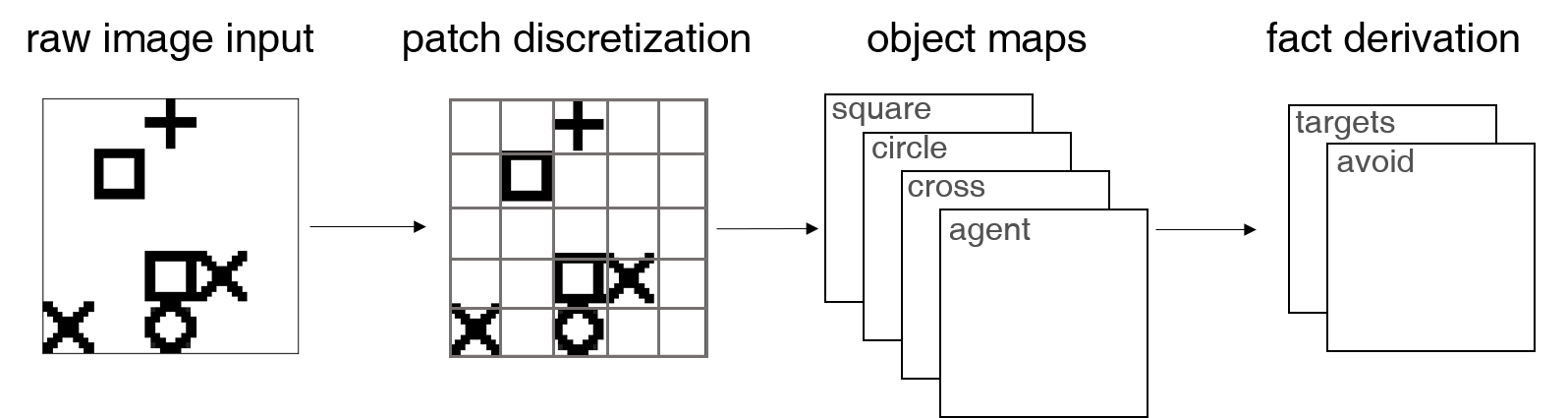}
    \caption{Priors on the grid game}
    \label{fig:priors_grid}
\end{figure}

The grounding of the type predicates is known and returns the corresponding channel of the object maps. We infer the grounding of the other predicates based on the background knowledge using Logic Tensor Networks (LTN) ~\cite{serafini_logic_2016}. The latent (unknown) relations are approximated using tensor networks. Training on the set of constraints -- the axioms --, LTN infers numerical groundings for $\mathrm{avoid},\mathrm{goto}$. This results in values for the provided facts on each point of the maps, represented in new maps (see also Figure \ref{fig:priors_grid}).

\section{Action Selection}
\begin{figure}
    \centering
    \includegraphics[width=1.\columnwidth]{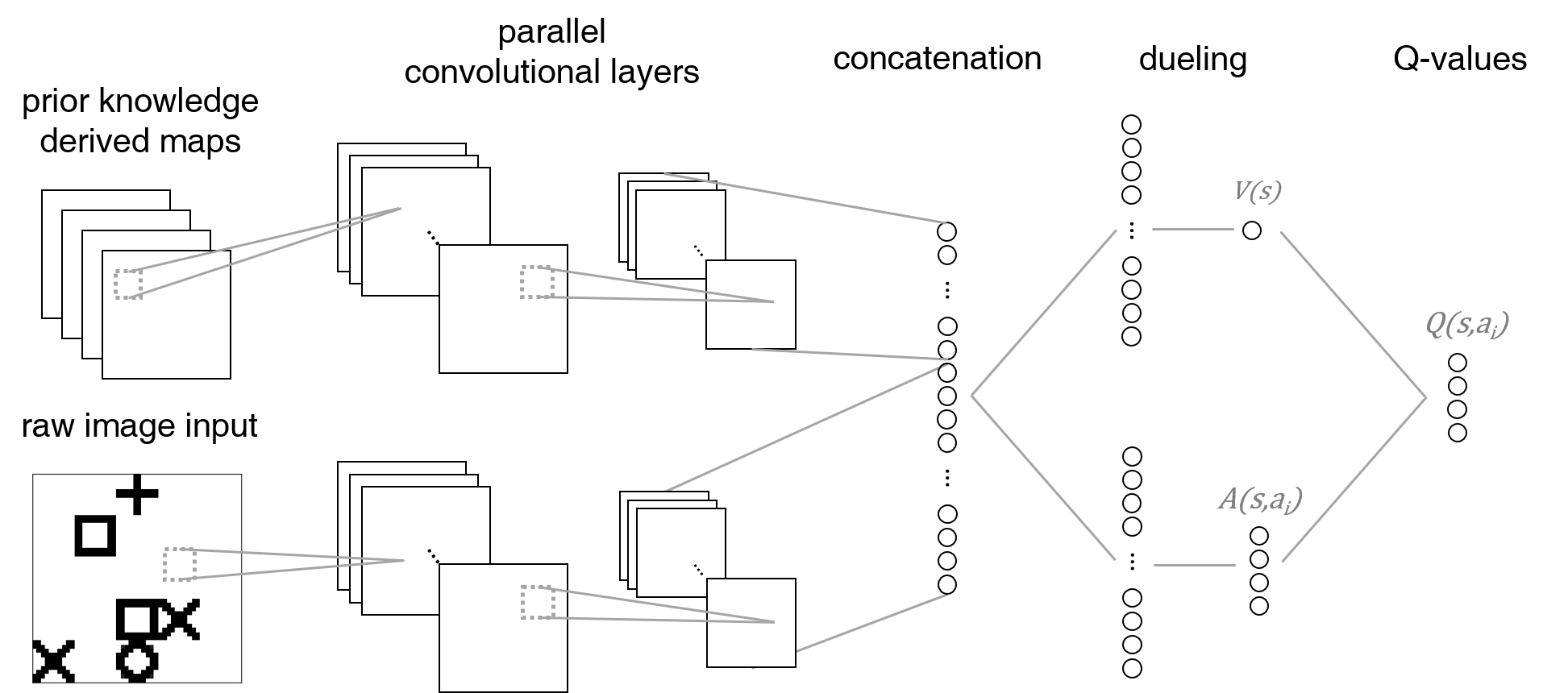}
    \caption{A Double Dueling architecture architecture with conjoint image and prior knowledge inputs.}
    \label{fig:DQN_architecture}
\end{figure}

Both the raw original input and the prior knowledge derived maps are fed into the action selection module.  Figure \ref{fig:DQN_architecture} shows our architecture applied to input/output streams. The original image input and the prior knowledge input are conjointly fed into the architecture. The action selection in our system is based on the Double Dueling Deep Q Network architecture~\cite{van_hasselt_double_2015} augmented by streams for processing symbolic and raw image information. The Double Dueling Deep Q Network is a variation of the Q-learning algorithm DQN~\cite{mnih_human-level_2015}, a value-based RL algorithm that learns to predict the expected discounted reward for state-action pairs $Q(s,a)$. DQN approximates the Q values using a deep neural network with various convolutional and fully-connected layers. The training uses an experience replay buffer for a more stable learning. 

The Double Dueling Deep Q Network architecture~\cite{hessel_rainbow:_2017} decouples the selection of the action from its evaluation~\cite{van_hasselt_double_2015}. Dueling approaches of value-based algorithms feature two stream of computation -- advantage depending on state-action and value depending on state only -- for a more robust estimation~\cite{wang_dueling_2015}. These improvements are compatible with our approach as we only investigate the input and not the action selection algorithm. 

Actions are selected $\epsilon$-greedily, \ie maximum valued action with probability $(1-\epsilon)$, random action with probability $\epsilon$. The rate of exploration $\epsilon$ is decreased over time.

\section{Results}
We train the framework on the experiments presented in Section \ref{sec:experiments}.
\begin{itemize}
    \item \textsc{Experiment I} tests an agent on different texture rendering settings of the game. We change setting every 50 epochs. It highlights the importance of symbolic abstraction and representing objects out of their visual context into types predicates.
    \item  \textsc{Experiment II} tests an agent on different scenarios of \emph{target type} and \emph{avoid type} objects. We change scenario every 50 epochs. It highlights the importance of providing facts on the objects to describe a task.
\end{itemize}

The background knowledge (object type classification and derived facts) is adjusted to the currently active scenario. We systematically vary the symbolic information injected into the agent architecture to show the effect. In our experiments, we test:
\begin{description}
\item[without priors] The agent only has the raw image as input for the action selection.
\item[with priors on object types] The results of the object recognition are joined to the raw image input.
\item[with priors on object types + facts] The results of the object recognition and fact derivation are joined to the raw image input.
\end{description}

Hyper-parameters of the RL algorithm such as $\epsilon$ can be optimized. We investigate two strategies for adjusting $\epsilon$.
\begin{description}
\item[$\epsilon$ reset] The decreasing exploration rate of the $\epsilon$-greedy action selection is reset at each new environment setting.
\item[$\epsilon$ not reset] The decreasing exploration rate of the $\epsilon$-greedy action selection is unchanged at each new environment setting.
\end{description}

The agents' network is trained every \num{2e2} timesteps. It is evaluated every \num{2e5} timesteps, defined as an {\it epoch}. Evaluations measure the collected rewards normalized by the potential maximum reward for a particular environment. Evaluations are averaged on 50 trajectories. The position of the agent and the position and number of objects are randomized at each trajectory. 

\begin{figure*}
    \centering
    \begin{subfigure}{.5\textwidth}
      \centering
      \includegraphics[width=0.9\textwidth]{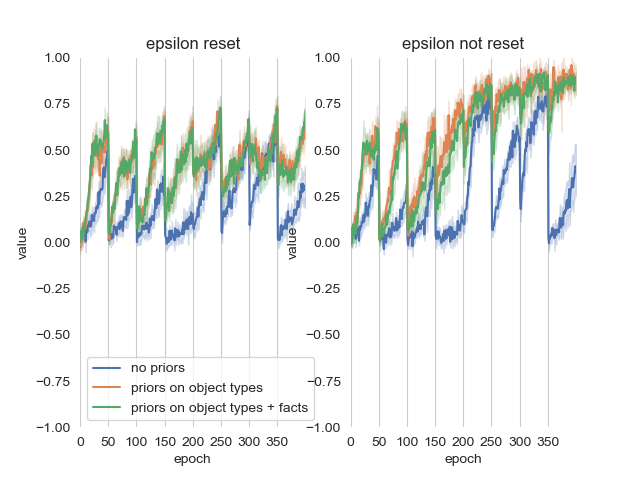}
      \caption{\textsc{Experiment I} - changing the rendering colors of the environment}
      \label{fig:results_exp1}
    \end{subfigure}%
    \begin{subfigure}{.5\textwidth}
      \centering
      \includegraphics[width=0.9\textwidth]{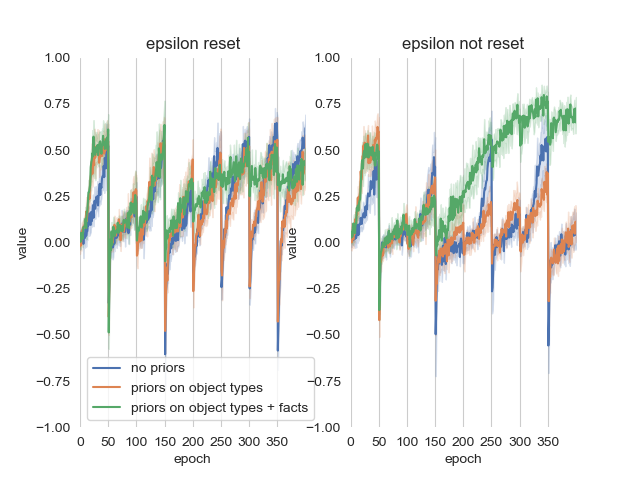}
      \caption{\textsc{Experiment II} - changing objects to target and to avoid}
      \label{fig:results_exp2}
    \end{subfigure}
    \caption{Average performance of agents with or without priors on object types and derived facts. The settings are changed every 50 epochs, the human-provided priors are updated accordingly. We consider experiments where agents reset (exploration oriented) or hold (exploitation oriented) their exploration parameters between each new setting.}
    \label{fig:results_exp}
\end{figure*}

Figure \ref{fig:results_exp1} and Figure \ref{fig:results_exp2} show results for 5 experimental runs per experiment. Also the 95\% confidence interval is plotted. We change scenarios and settings every 50 epochs during 400 epochs, for a total of 8 changes. \textsc{Experiment I} proves that the agent successfully leverages the knowledge on object types through time. \textsc{Experiment II} proves that the agent successfully leverages the knowledge on facts through time.

\subsection{Results Experiment I}
Figure \ref{fig:results_exp1} shows results for Experiment I. The baseline \emph{no prior} condition shows that every time the scenario changes, the algorithm is relearning the task. This behavior does not actually depend on whether $\epsilon$ is reset or not. 

If we check the performance of \emph{priors on object types}, then we can see 2 trends vs the baseline. 1) The learning seems to be much faster. 2) The drop in performance at each scenario change becomes smaller and smaller. In other words the system still has to relearn the task initially but over time becomes more and more immune to scenario/color changes. The system has learned to rely on object type information to become performant even when the color of the object changes. In summary, object priors help the system to learn faster and to be able to transfer task performance between different scenarios. Notice that the effect is more pronounced when $\epsilon$ is not reset. As opposed to the baseline, $\epsilon$ reset actually matters as it will favor exploration over exploitation. So because the system has learned to generalize across scenarios, $\epsilon$ should not be reset.

Lastly, if we check the \emph{priors on object types and facts} then we can see almost no difference to the performance of \emph{priors on object types}. This makes sense as only the color of objects change and not their avoidance/collection semantics. Consequently, knowledge about the object semantics is not relevant for learning to generalize over Experiment I scenario changes.

\subsection{Results Experiment II}
Figure \ref{fig:results_exp2} shows results for Experiment II. Similar to Experiment I, the graph shows that in the baseline \emph{no prior} condition, the algorithm is relearning the task every time the scenario changes. While there is some impact of whether $\epsilon$ is reset or not, overall the system is mostly relearning the task (unless there is some overlap between succeeding scenarios).

If we check the performance of \emph{priors on object types} condition, then we can see that it mirrors the baseline condition. Remember that in this experiment a circle can be an object to avoid in one scenario and is an object to collect in another. Consequently, knowledge about object types does not help to learn faster or be able to transfer knowledge from one scenario to another.

If we check the \emph{priors on object types and facts} condition, then we can see 2 trends vs the baseline (and the \emph{priors on object types} condition). 1) The learning seems to be much faster. 2) The drop in performance at each scenario change becomes smaller and smaller. In other words the system still has to relearn the task initially but over time becomes more and more immune to scenario changes. The system has learned to rely on derived facts to become performant even when the semantics of an object with respect to the task changes. In summary, object priors help the system to learn faster and to be able to transfer task performance between different scenarios.

In this last condition resetting $\epsilon$ does have impact on the general trend in learning. Resetting $\epsilon$ hurts the baseline system and the system using only object type priors, but aids the system using all background knowledge. Overall performance is higher and learning is faster. This makes sense, because if the system has learned to transfer across scenario changes, then resetting $\epsilon$ is undesirable. On the other hand, resetting $\epsilon$ does aid the baseline and object type prior systems. For systems that do not learn to transfer, resetting $\epsilon$ at least allows them to learn each scenario over and over again.

\section{Conclusion}
This paper discussed a new approach to injecting prior knowledge conjointly with original raw input into reinforcement learning algorithms. By grounding these priors in predicates, we showed how symbolic semantics on objects are useful for transfer learning. As proof-of-concept, we demonstrated the architecture in a simple grid world. The experiments show that symbolic abstractions can help to solve tasks across various scenarios without relearning the decision module. We demonstrated how agents learn to leverage the appropriate knowledge for a particular task and learn to select and exploit information from prior knowledge sources.

Further work should test this approach in more complex environments with more or less human-provided information. Here, we investigated priors as a state expansion method for transfer learning. We plan to explore other ways to use the priors in a RL architecture, such as a policy transfer method or reward function indicator. We also plan to further investigate the impact of exploration strategies on the system, to prevent misguidance from inaccurate priors. More elaborate exploration frameworks have recently been investigated in the literature~\cite{oudeyer_intrinsic_2009,pathak_curiosity-driven_2017}. Such algorithms might become essential when using prior knowledge in order to tradeoff exploration and exploitation at the correct time in the experiment.

\bibliographystyle{named}
\bibliography{ijcai19}

\end{document}